\documentclass[runningheads]{llncs}
\usepackage{graphicx}
\graphicspath{{./}}
\usepackage{amsmath}
\usepackage{svg}
\usepackage{caption}
\usepackage{subcaption}
\usepackage{url}
\usepackage[hyperfootnotes=false]{hyperref}
\usepackage{footnote}
\usepackage[symbol]{footmisc}

\begin{document}
\title{LineFormer: Rethinking Line Chart Data Extraction as Instance Segmentation}
\titlerunning{LineFormer}
%
\author{Jay Lal\thanks{Corresponding Author} \and
Aditya Mitkari\thanks{Joint Second Authorship} \and
Mahesh Bhosale\protect\footnotemark[2] \and 
David Doermann}

\authorrunning{J. Lal et al.}
%

\institute{Artificial Intelligence Innovation Lab (A2IL), \\
Department of Computer Science and Engineering, \\
University at Buffalo, Buffalo, NY, USA\\
\email{\{jayashok, amitkari, mbhosale, doermann\}@buffalo.edu}
}
%
\maketitle              
\begin{abstract}
Data extraction from line-chart images is an essential component of the automated document understanding process, as line charts are a ubiquitous data visualization format. However, the amount of visual and structural variations in multi-line graphs makes them particularly challenging for automated parsing. Existing works, however, are not robust to all these variations, either taking an all-chart unified approach or relying on auxiliary information such as legends for line data extraction. In this work, we propose LineFormer, a robust approach to line data extraction using instance segmentation. We achieve state-of-the-art performance on several benchmark synthetic and real chart datasets. Our implementation is available at \url{https://github.com/TheJaeLal/LineFormer}.

\end{abstract}
\section{Introduction}
Automated parsing of chart images has been of interest to the research community for some time now. Some of the downstream applications include the generation of new visualizations~\cite{chartredsign}, textual summarization~\cite{demir2012summarizing}, retrieval~\cite{chartretrieval} and visual question answering~\cite{masry_chartqa_2022}. Although the related text and image recognition fields have matured, research on understanding figures, such as charts, is still developing. Most charts, such as bar, pie, lines, scatter, etc., typically contain textual elements (title, axis labels, legend labels) and graphical elements (axis, tick marks, graph/plot). Like OCR is often a precursor to any document understanding task, data extraction from these chart images is a critical component of the figure analysis process.

\begin{figure}[htp]
    \centering
        \includegraphics[width=1\textwidth]{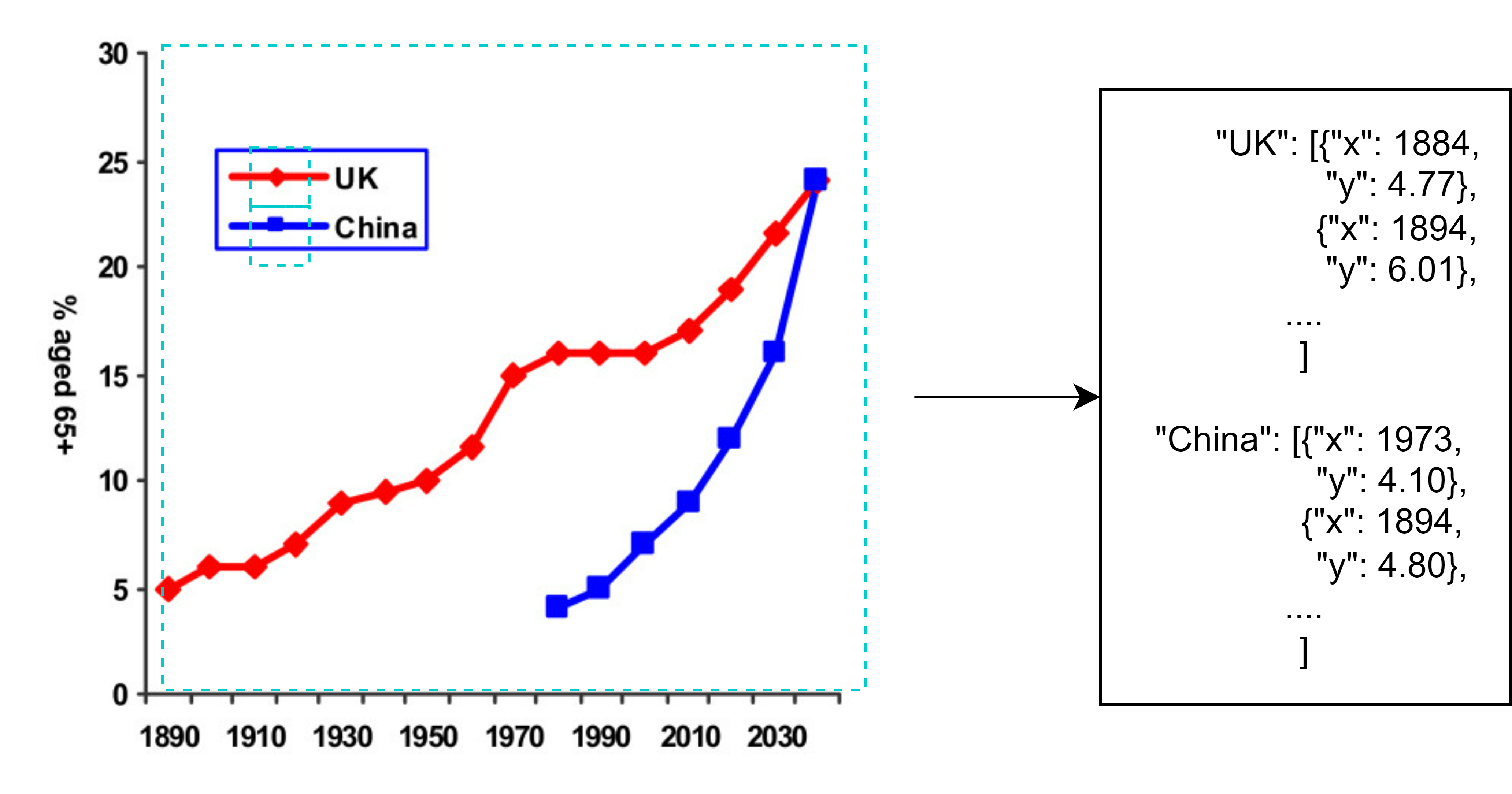}
        \caption{Data Extraction}
        \label{fig:AuxTask}
\end{figure}

\begin{figure}[!htp]
    \centering
     \begin{subfigure}[b]{0.40\textwidth}
         \centering
         \includegraphics[width=\textwidth]{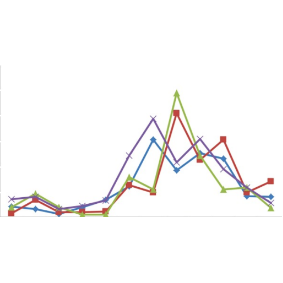}
         \caption{Original Image}
     \end{subfigure}
     \begin{subfigure}[b]{0.40\textwidth}
         \centering
         \includegraphics[width=\textwidth]{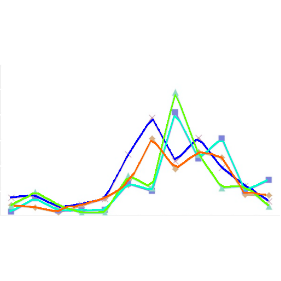}
         \caption{LineFormer}
     \end{subfigure}
     \begin{subfigure}[b]{0.40\textwidth}
         \centering
         \includegraphics[width=\textwidth]{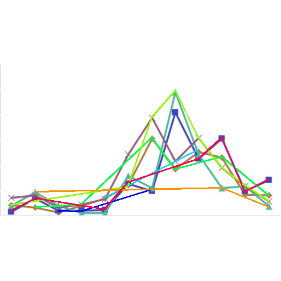}
         \caption{ChartOCR~\cite{luo_chartocr_2021}}
     \end{subfigure}
     \begin{subfigure}[b]{0.40\textwidth}
         \centering
         \includegraphics[width=\textwidth]{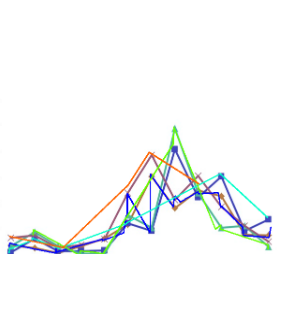}
         \caption{LineEx~\cite{p_lineex_2023}}
     \end{subfigure}
        \caption{Qualitative analysis: PMC samples}
        \label{fig:Qual_PMC}
\end{figure}

\begin{figure}[!htp]
    \centering
     \begin{subfigure}[b]{0.22\textwidth}
         \centering
         \includegraphics[width=\textwidth]{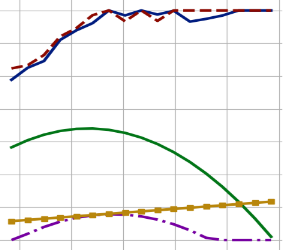}
     \end{subfigure}
     \begin{subfigure}[b]{0.22\textwidth}
         \centering
         \includegraphics[width=\textwidth]{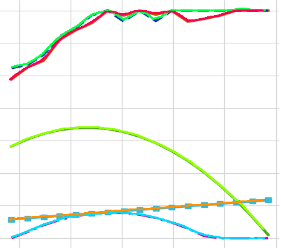}
     \end{subfigure}
     \begin{subfigure}[b]{0.22\textwidth}
         \centering
         \includegraphics[width=\textwidth]{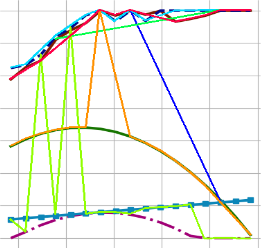}
     \end{subfigure}
     \begin{subfigure}[b]{0.22\textwidth}
         \centering
         \includegraphics[width=\textwidth]{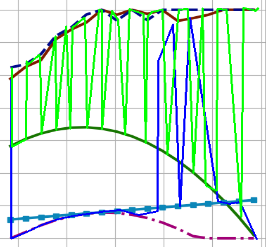}
     \end{subfigure}
     \begin{subfigure}[b]{0.22\textwidth}
         \centering
         \includegraphics[width=\textwidth]{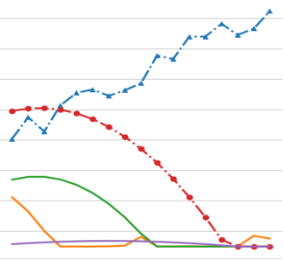}
         \caption{Original Image}
     \end{subfigure}
     \begin{subfigure}[b]{0.22\textwidth}
         \centering
         \includegraphics[width=\textwidth]{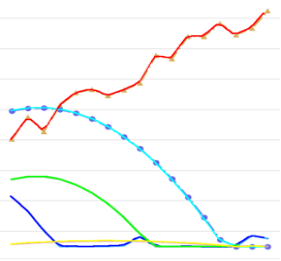}
         \caption{LineFormer}
     \end{subfigure}
     \begin{subfigure}[b]{0.22\textwidth}
         \centering
         \includegraphics[width=\textwidth]{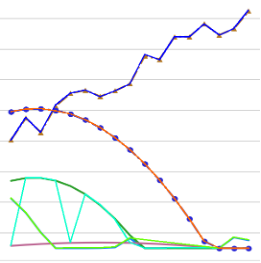}
         \caption{ChartOCR~\cite{luo_chartocr_2021}}
     \end{subfigure}
     \begin{subfigure}[b]{0.22\textwidth}
         \centering
         \includegraphics[width=\textwidth]{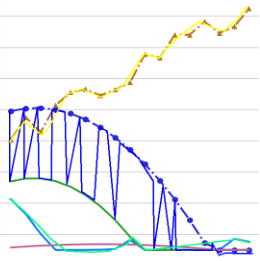}
         \caption{LineEx~\cite{p_lineex_2023}}
     \end{subfigure}
        \caption{Qualitative analysis: LineEx samples}
        \label{fig:Qual_Linex}

\end{figure}

For this study, we have chosen to focus on the analysis of line charts. Despite their widespread use, line charts are among the most difficult to parse accurately (see Figs. \ref{fig:Qual_PMC},\ref{fig:Qual_Linex}). A chart data extraction pipeline typically involves many auxiliary tasks, such as classifying the chart type and detecting text, axes, and legend components before data extraction (see Fig. \ref{fig:AuxTask}). Here, we concentrate on the final task of data series extraction, that is, given a chart image, we output an ordered set of points x,y (image coordinates), representing the line data points. Our solution can be easily extended to recover the tabular data originally used to generate the chart figure by mapping the extracted series to given axis units.

Recent works\cite{luo_chartocr_2021,ma_towards_2021} show that box and point detectors have achieved reasonable accuracy in extracting data from bar, scatter, and pie charts, as they often have fewer structural or visual variations. However, line charts pose additional challenges because of the following. 

\begin{itemize}
  \item Plethora of visual variations in terms of line styles, markers, colors, thickness, and presence of background grid lines.
  \item Structural complexities, such as crossings, occlusions, and crowding. (Fig. [\ref{fig:struct_complex}])
  \item Absence of semantic information, such as legends, to differentiate lines with a similar style or color.  
\end{itemize}

Although some recent works propose specialized solutions for lines, they still fail to address some of the above problems (see Fig. \ref{fig:Qual_PMC}, \ref{fig:Qual_Linex}). In this work, we propose a robust method for extracting data from line charts by formulating it as an instance segmentation problem. Our main contributions can be summarized as follows.

\begin{itemize}
    \item We highlight the unique challenges in extracting data from multi-line charts.
    \item We propose LineFormer, a simple, yet robust approach that addresses the challenges above by reformulating the problem as instance segmentation. 
    \item We demonstrate the effectiveness of our system by evaluating benchmark datasets. We show that it achieves state-of-the-art results, especially on real chart datasets, and significantly improves previous work.
\end{itemize}

We also open-source our implementation for reproducibility.

\section{Related Work}

\subsection{Chart Analysis}
Analyzing data from chart images has been approached from multiple perspectives. Chart-to-text~\cite{chartsummary} tried to generate natural language summaries of charts using transformer models. Direct Visual Question Answering  (VQA) over charts has also been another popular area of research~\cite{masry_chartqa_2022}. Many high-level understanding tasks can be performed more accurately if the corresponding tabular data from the chart have been extracted, making the extraction of chart data an essential component for chart analysis.

\subsection{Line Data Extraction}
\subsubsection{Traditional Approaches}
Earlier work on data extraction from line charts used image-processing-based techniques and heuristics. These include color-based segmentation \cite{74}, \cite{115}, or tracing using connecting line segments~\cite{46} or pixels~\cite{110}. Patch-based methods capable of recognizing dashed lines have also been used~\cite{58}. These methods work well for simple cases; however, their performance deteriorates as the number of lines increases, adding more visual and structural variations.
\subsubsection{Keypoint Based}
More recently, deep learning-based methods for extracting chart data have gained popularity. Most of these approaches attempt to detect lines through a set of predicted keypoints and then group those belonging to the same line to differentiate for multi-line graphs. ChartOCR\cite{luo_chartocr_2021} used a CornerNet-based keypoint prediction model and push-pull loss to train keypoint embeddings for grouping. Lenovo\cite{ma_towards_2021} used an FPN-based segmentation network to detect line keypoints. Recently, LineEX\cite{p_lineex_2023} extended ChartOCR by replacing CornerNet with Vision Transformer and forming lines by matching keypoint image patches with legend patches. A common problem with these keypoint-oriented approaches is that predicted points do not align with the ground-truth line precisely. Furthermore, robust keypoint association requires aggregating contextual information that is difficult to obtain with only low-level features or push-pull-based embeddings.
Some segmentation-based tracking approaches have also been proposed, treating the line as a graph with line pixels as nodes and connections as edges~\cite{kato_parsing_2022,siegel_figureseer_2016}. Lines are traced by establishing the presence of these edges using low-level features such as color, style, etc., or by matching with the legend to calculate the connection cost. The cost is minimized using Dynamic Programming\cite{siegel_figureseer_2016} or Linear Programming\cite {kato_parsing_2022} to obtain the final line tracking results.

We refer the reader to \cite{chartsurvey2020,CHART-2020,CHART-2022} for a more detailed chart comprehension and analysis survey.

\subsection{Instance Segmentation}
The goal of instance segmentation is to output a pixel-wise mask for each class instance. Earlier approaches, for instance segmentation such as MaskRCNN~\cite{maskrcnn}, were derived from top-down object detection techniques. However, when segmenting objects with complex shapes and overlapping instances, bottom-up approaches have performed better~\cite{instseg}. Bottom-up approaches, for instance segmentation, extend semantic segmentation networks by using a discriminative loss~\cite{instseg} to cluster pixels belonging to the same instance. More recently, transformer-based architectures have achieved state-of-the-art results on instance segmentation tasks~\cite{videoinstseg,mask2former}. These architectures are typically trained using a set prediction objective and output a fixed set of instance masks by processing object queries.

\section{Motivation}

\subsection{Line Extraction Challenges}\label{line_challenges}
\begin{figure}[htp]
    \centering
    \includegraphics[width=0.6\textwidth]{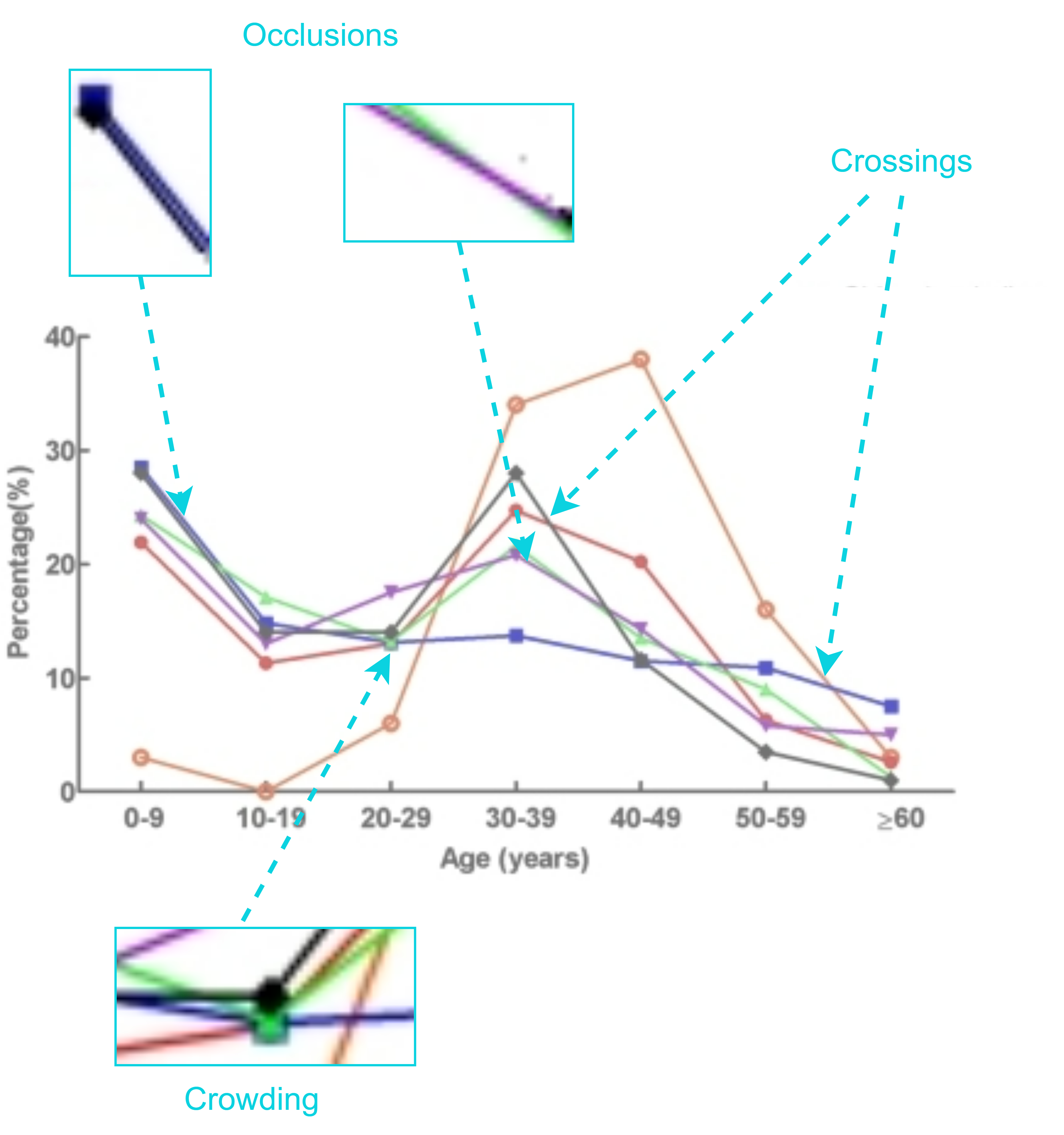}
    \caption{Line - Structural Complexity Patterns}
    \label{fig:struct_complex}
\end{figure}

As mentioned, line charts, especially multiline graphs, suffer from several structural complexities. We identify the three most common structural patterns: crossings, occlusions, and crowding. Fig. \ref{fig:struct_complex} shows their examples.

In each of these cases, the visual attributes of the line segments, such as color, style, and markers, are either obscured or blended with adjacent lines. Thus, approaches that primarily rely on low-level image features, such as color, gradients, texture, etc., often fall short. Furthermore, most keypoint-based line extractors \cite{luo_chartocr_2021,p_lineex_2023} tend to suffer from two common issues: a.) The inability to predict distinct keypoint for each line at crossings and occlusions. b.) The subsequent keypoint grouping step suffers as it attempts to extract features from an already occluded local image patch. (See Figs. \ref{fig:Qual_PMC},\ref{fig:Qual_Linex}).

Recent work ~\cite{kato_parsing_2022} attempts to address occlusion by modeling an explicit optimization constraint. However, the tracking remains dependent on low-level features and proximity heuristics, which could limit its robustness. It is worth noting that, in such cases, humans can gather contextual information from surrounding areas and fill the gaps.

\subsection{Line As Pixel Mask}
Due to the shortcomings in the keypoint-based method (Section\ref{line_challenges}), we adopt a different approach inspired by the task of lane detection in autonomous driving systems. In this context, instance segmentation techniques have effectively detected different lanes from road images~\cite{lanedetection2018}. Similarly to lanes, we consider every line in the chart as a distinct instance of the class ‘line’ and predict a binary pixel mask for each such instance. Thus, rather than treating a line as a set of arbitrary keypoints, we treat it as a group of pixels and demonstrate that per-pixel line instance segmentation is a more robust learning strategy than keypoint detection for line charts. Moreover, this approach does not require an explicit keypoint-grouping mechanism since all pixels (or points) segmented in a mask belong to the same line. 

\section{Approach}
\begin{figure}[htp]
    \centering
    \includegraphics[width=1\textwidth]{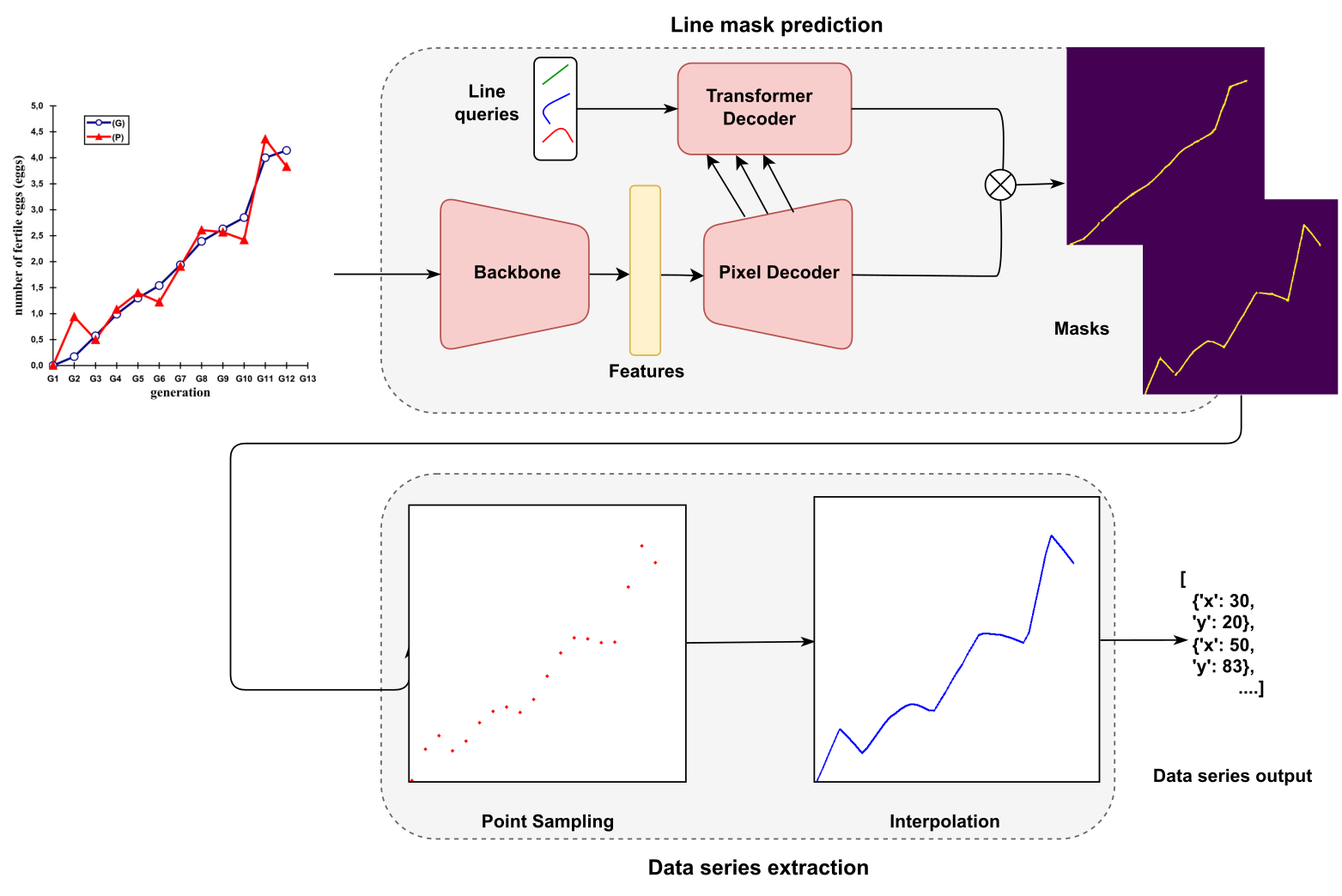}
    \caption{LineFormer System}
    \label{fig:sys_overview}
\end{figure}

To address the problem of occlusion and crowding, we adopt an instance segmentation model based on the encoder-decoder transformer~\cite{mask2former}, which uses a masked-attention pixel decoder capable of providing the visual context necessary to detect occluded lines. Our proposed LineFormer system (Fig.\ref{fig:sys_overview}) is simple and consists of two modules, Mask Prediction and Data Series Extraction.

\subsection{Line Mask Prediction}
We adopt a state-of-the-art universal segmentation architecture to predict line instances from charts~\cite{mask2former}. The input line chart image $I_{H*W}$ is processed by a transformer model, an encoder-decoder network. The encoder extracts multi-level features $f$ from the input image $I$, followed by a pixel decoder that up-samples the extracted features $f$ and outputs a 2D per pixel embedding $P_{C*H*W}$. A transformer decoder attends to the intermediate layer outputs of the pixel decoder and predicts a mask embedding vector $M_{C*N}$, where $N$ is equal to the number of line queries. Finally, the dot product of $P$ and $M$ gives an independent per-pixel mask $E_{H*W*N}$.

\subsection{Data Series Extraction}
Once the predicted line masks $E$ for an input image are obtained, we find the start and end x-values of the line $x_{range}$ = [$x_{start}$, $x_{end}$]. Within $x_{range}$, we sample foreground points ($x_i$, $y_i$) at a regular interval $\delta_{x}$. The smaller the interval, $\delta_{x}$, the more precisely the line can be reconstructed. We used linear interpolation on the initial sampled points to address any gaps or breaks in the predicted line masks. The extracted ($x$,$y$) points can be scaled to obtain the corresponding data values if the information on the ground truth axis is available.

\section{Implementation Details}
We adopt the architecture and hyperparameter settings from \cite{mask2former}. For the backbone, we use SwinTransformer-tiny~\cite{swint} to balance accuracy and inference speed. To ensure a manageable number of line predictions, we set the number of line queries to 100. As the transformer decoder is trained end-to-end on a set prediction objective, the loss is a linear combination of classification loss and mask prediction loss, with weights 1 and 5, respectively. The latter is the sum of dice loss~\cite{diceloss} and cross-entropy loss. All experiments are performed using the MMDetection~\cite{mmdetection} framework based on PyTorch.

\subsection{Generating Ground Truth Line Masks}
Training LineFormer requires instance masks as labels. Thus, for a chart image, we generate a separate Boolean ground-truth mask for each line instance. As mentioned earlier, these line masks may have overlapping pixels that reflect the natural structure in the corresponding chart images. Since all datasets for line data extraction provide annotations in the form of x,y points, we need to convert these points into pixel line masks. For this, we use the Bresenham algorithm~\cite{bresenham} from computer graphics to generate continuous line masks similar to how the lines would have been rendered in the corresponding plot image. By doing this, we can maintain a 1-1 pixel correspondence between the input image and the output pixel, making the prediction task easier and stabilizing the training. We fix the thickness of lines in ground-truth masks to three pixels, which is empirically found to work best for all the datasets.

\section{Experiments}

\subsection{Baselines}
To test our hypothesis, we performed multiple experiments on several real and synthetic chart datasets. We compare LineFormer with keypoint-based approaches as baselines, including line-specific LineEX~\cite{p_lineex_2023} and unified approaches ChartOCR~\cite{luo_chartocr_2021} and Lenovo~\cite{ma_towards_2021}. We take their reported results for the latter, as they report on the same CHART-Info challenge\cite{CHART-2020} metrics that we use. For ChartOCR~\cite{luo_chartocr_2021} and LineEX~\cite{p_lineex_2023}, we use their publicly available pre-trained models to evaluate on the same competition metrics. We cannot compare with the Linear Programming method~\cite{kato_parsing_2022}  due to the inaccessibility of its implementation and because its reported results are based on a different metric.

\subsection{Datasets}
Training data-extraction models require a large amount of data. Furthermore, no single dataset has a sufficient number of annotated samples in terms of quantity and diversity to train these models. Thus, following suit with recent works \cite{kato_parsing_2022,ma_towards_2021,p_lineex_2023}, we combine multiple datasets for model training.

\subsubsection{AdobeSynth19}:
This dataset was released from the ICDAR-19 Competition on Harvesting Raw Tables from Infographics 2019 (CHART-Info-19)\cite{CHART-2019}. It contains samples of different chart types and has nearly 38,000 synthetic line images generated using matplotlib and annotations for data extraction and other auxiliary tasks. We use the competition-provided train test split.

\subsubsection{UB-PMC22}:
This is the extended version of the UB-PMC-2020 dataset published by CHART-Info-2020~\cite{CHART-2022}. It consists of real chart samples obtained from PubMed Central that were manually annotated for several tasks, including data extraction. Again, we use the provided train test split, which contains about 1500 line chart samples with data extraction annotation in the training set and 158 in the testing set. It is one of the most varied and challenging real-world chart datasets publicly available for data extraction.

\subsubsection{LineEX}:
This is another synthetic dataset released with LineEX\cite{p_lineex_2023} work, which contains more variations in structural characteristics - line shape, crossings, etc., and visual characteristics - line style and color. It was generated using matplotlib and is the largest synthetic dataset available for line chart analysis. We only use a subset, 40K samples from their train set and 10K samples from the test set for evaluation.

\section{Evaluation}
The particular choice of metric to evaluate line data extraction performance varies amongst existing works, yet they all share a common theme. The idea is to calculate the difference between the y values for the predicted and ground-truth line series and to aggregate that difference to generate a match score. ChartOCR~\cite{luo_chartocr_2021} averages these normalized y value differences across all ground truth x,y points, while Figureseer~\cite{siegel_figureseer_2016} and Linear Programming \cite{kato_parsing_2022} binarize the point-wise difference by taking an error threshold of 2\% to consider a point as correctly predicted. Here, we use the former as it is more precise in capturing the more minor deviations in line point prediction. Furthermore, the CHART-Info challenge formalized the same, proposing the Visual Element Detection Score and the Data Extraction Score, also referred to as task-6a and task-6b metrics \cite{CHART-2020}.

\subsection{Pairwise Line Similarity}\label{pairwise_similarity}
Formally, to compare a predicted and ground truth data series, a similarity score is calculated by aggregating the absolute difference between the predicted value $\Bar{y_i}$ and the ground truth value $y_i$, for each ground truth data point $(x_i, y_i)$. The predicted value $\Bar{y_i}$ is linearly interpolated if absent.

\subsection{Factoring Multi-Line Predictions}
Most charts contain multiple lines. Thus, a complete evaluation requires a mapping between ground-truth and predicted lines. The assignment is done by computing the pairwise similarity scores (as shown in \ref{pairwise_similarity}) between each predicted and ground truth line and then performing a bipartite assignment that maximizes the average pairwise score. This is similar to the mean average precision (mAP) calculation used to evaluate Object Detection performance.

Let $N_p$ and $N_g$ be the number of lines in the predicted and ground-truth data series.

We define K as the number of columns in the pairwise similarity matrix $S_{ij}$ and the bipartite assignment matrix $X_{ij}$, where $i$ ranges $[1, N_{p}]$ and $j$ ranges $[1, N_{g}]$

\begin{align*}
& K = \begin{cases}
N_g \hspace{90pt}\text{ for Visual Element Detection Score(task 6a) }\\
max(N_g, N_p) \hspace{46pt}\text{ for Data Extraction Score (task 6b) }
\end{cases} \\
& S_{ij} = \begin{cases}
    0 \hspace{95pt}\text{ if } j > N_g \\
    Similarity(P_i, G_j) \hspace{18pt}\text{ otherwise} 
\end{cases} \\ 
& X_{ij} = \begin{cases}
    1 \hspace{95pt}\text{ if } P_{i} \text{ matched with } G_{j} \\
    0 \hspace{97pt} \text{otherwise}
\end{cases}
\end{align*}


The final score is calculated by optimizing the average pairwise similarity score under the one-to-one bipartite assignment constraint.
\[
score = \frac{1}{K}*\underset{X}{max}\sum_{i}^{}\sum_{j}^{}S_{ij}X_{ij} \hspace{10pt}s.t\hspace{5pt} \sum_{i}X_j = 1 \hspace{7pt} and \hspace{7pt} \sum_{j}X_i = 1
\]

The difference between task-6a and task-6b scores is that the former only measures line extraction recall and hence doesn't consider the false positives (extraneous line predictions). In task-6b, however, a score of ‘0’ is assigned for each predicted data series that hasn’t been matched with a corresponding ground truth, along with an increase in value of denominator $K$ for normalizing the score.

\section{Results}

\subsection{Quantitative Analysis}
\begin{table}
\caption{LineFormer Quantitative Evaluation}\label{tab1}
\begin{minipage}{\textwidth}
\begin{tabular}{|l|cc|cc|cc|}
\hline
Dataset            & \multicolumn{2}{c|}{AdobeSynth19}  & \multicolumn{2}{c|}{UB-PMC22}        & \multicolumn{2}{c|}{LineEX}        \\ \hline
Work \textbackslash Metric &
  \multicolumn{1}{l|}{\begin{tabular}[c]{@{}l@{}}Visual \\ Element \\ Detection\footnote{task-6a from CHART-Info challenge~\cite{CHART-2020}}\end{tabular}} &
  \multicolumn{1}{l|}{\begin{tabular}[c]{@{}l@{}}Data \\ Extraction\footnote{task-6b data score from CHART-Info challenge}\end{tabular}} &
  \multicolumn{1}{l|}{\begin{tabular}[c]{@{}l@{}}Visual \\ Element\\ Detection\end{tabular}} &
  \multicolumn{1}{l|}{\begin{tabular}[c]{@{}l@{}}Data \\ Extraction\end{tabular}} &
  \multicolumn{1}{l|}{\begin{tabular}[c]{@{}l@{}}Visual \\Element\\ Detection\end{tabular}} &
  \multicolumn{1}{l|}{\begin{tabular}[c]{@{}l@{}}Data\\ Extraction\end{tabular}} \\ \hline
ChartOCR~\cite{luo_chartocr_2021}           & \multicolumn{1}{c|}{84.67} & 55    & \multicolumn{1}{c|}{83.89}  & 72.9   & \multicolumn{1}{c|}{86.47} & 78.25 \\ \hline
Lenovo\footnote{Implementation not public, hence only reported numbers on AdobeSynth and UB PMC}~\cite{ma_towards_2021} & \multicolumn{1}{c|}{\textbf{99.29}} & \textbf{98.81} & \multicolumn{1}{c|}{84.03}  & 67.01  & \multicolumn{1}{c|}{-}     & -     \\ \hline
LineEX~\cite{p_lineex_2023} & \multicolumn{1}{c|}{82.52} & 81.97 & \multicolumn{1}{c|}{50.23\footnote{Ignoring samples without legend, as LineEX doesn't support them}} & 47.03 & \multicolumn{1}{c|}{71.13} & 71.08 \\ \hline
Lineformer (Ours)  & \multicolumn{1}{c|}{97.51} & 97.02 & \multicolumn{1}{c|}{\textbf{93.1}}   & \textbf{88.25}  & \multicolumn{1}{c|}{\textbf{99.20}} & \textbf{97.57} \\ \hline
\end{tabular}
\end{minipage}
\end{table}

The performance of various systems for line data extraction on visual element detection and data extraction is shown in Table \ref{tab1}. It can be observed that the UB-PMC data set proves to be the most challenging of all, as it is diverse and composed of real charts from scientific journals. LineFormer demonstrates state-of-the-art results on most real and synthetic datasets. Furthermore, our results are consistent across the two evaluation metrics: Visual Element Detection (task-6a) and Data Extraction (task-6b). This stability is not reflected in existing work, as they exhibit substantial drops in performance. This discrepancy can be attributed to the fact that Task-6b penalizes false positives, indicating that LineFormer has a considerably higher precision rate. A key highlight of LineFormer is that it is robust across datasets and shows significant improvement on UB-PMC real-world chart data.

\subsection{Qualitative Analysis}

The qualitative comparison of LineFormer with existing models is carried out by examining line predictions on selected samples from the dataset. This analysis focuses on scenarios that involve crossings, occlusions, and crowding to assess the ability of models to handle such complex situations. The results are presented in Fig [\ref{fig:Qual_PMC},\ref{fig:Qual_Linex}], which clearly illustrate the difficulties faced by existing approaches as the number of lines and the complexity of the chart increases. LineFormer performs significantly better, keeping track of the line even in occlusions and crowding.

\subsection{Ablation Study}
\begin{table}
\centering
\caption{Performance of different backbones on UB-PMC data}\label{tab-ablation}
\begin{tabular}{|l|c|c|}
\hline
Backbone & PMC - 6a & \multicolumn{1}{l|}{PMC - 6b} \\ \hline
Swin-T~\cite{swint} & 93.1 & 88.25 \\ \hline
ResNet50~\cite{resnet} & 93.17 & 87.21 \\ \hline
Swin-S~\cite{swint} & \textbf{93.19} & 88.51 \\ \hline
ResNet101~\cite{resnet} & 93.1 & \textbf{90.08} \\ \hline
\end{tabular}
\end{table}

We conducted an ablation study to understand the impact of different backbones on line extraction performance. For simplicity, we stick to the UB-PMC dataset, which has the most diverse set of real-world line charts. Table \ref{tab-ablation} shows the results of the study. We observe that performance remains mostly stable across most backbones, with a general trend towards an increase in 6b scores with an increase in the number of parameters. Furthermore, it should be noted that even the smallest backbone, SwinTransformer-Tiny~\cite{swint} (Swin-T), surpasses the current keypoint-based state of the art (see Table~\ref{tab1}) on the UB-PMC dataset by a large margin. 
\\ \\ 
The ablation study demonstrates that the choice of backbone has a relatively minor impact on the overall performance of line extraction. This suggests that LineFormer is robust to different backbone architectures, and the main advantage is derived from the overall framework and the instance segmentation approach.


\section{Discussion}

After extracting the line data series, each line must be associated with its corresponding label in the legend. To accomplish this task, a Siamese network can match the line embeddings with those of the patches in the provided legend.

\begin{figure}[!htp]
    \centering
     \begin{subfigure}[b]{0.3\textwidth}
         \centering
         \includegraphics[width=\textwidth]{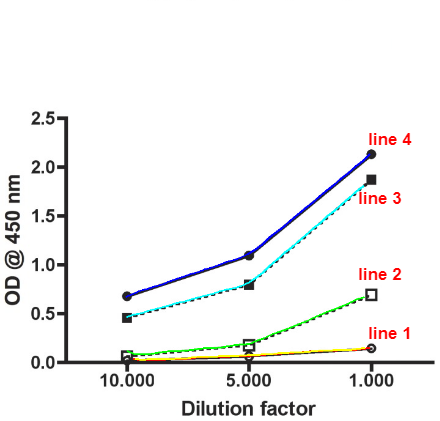}
         \caption{annotated image}
         \label{fig:limit_annot_img}
     \end{subfigure}
     \hfill
     \begin{subfigure}[b]{0.3\textwidth}
         \centering
         \includegraphics[width=\textwidth]{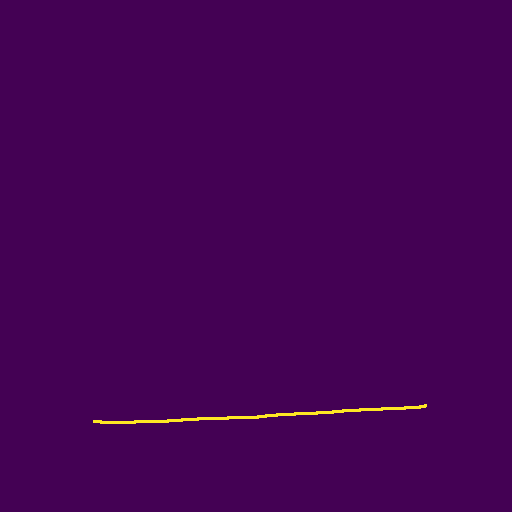}
         \caption{mask line 1}
         \label{fig:mask_line_1}
     \end{subfigure}
     \hfill
     \begin{subfigure}[b]{0.3\textwidth}
         \centering
         \includegraphics[width=\textwidth]{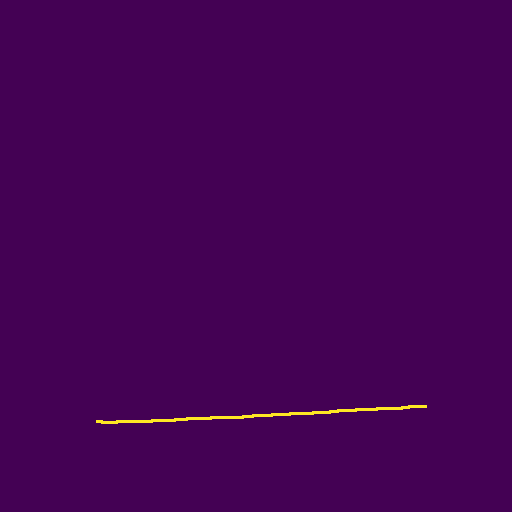}
         \caption{Duplicate mask line 1}
         \label{fig:mask_line_1_d}
     \end{subfigure}
     \hfill
     \begin{subfigure}[b]{0.3\textwidth}
         \centering
         \includegraphics[width=\textwidth]{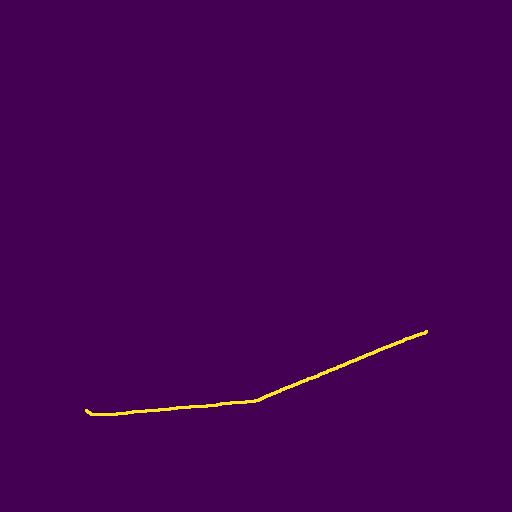}
         \caption{mask line 2}
         \label{fig:mask_line_2}
     \end{subfigure}
     \hfill
     \begin{subfigure}[b]{0.3\textwidth}
         \centering
         \includegraphics[width=\textwidth]{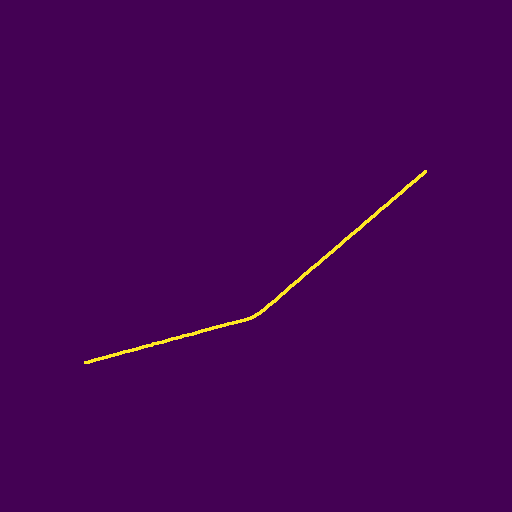}
         \caption{mask line 3}
         \label{fig:mask_line_3}
     \end{subfigure}
     \hfill
     \begin{subfigure}[b]{0.3\textwidth}
         \centering
         \includegraphics[width=\textwidth]{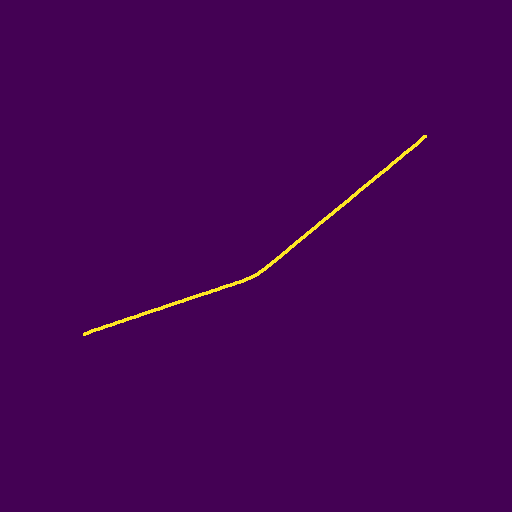}
         \caption{mask line 4}
         \label{fig:mask_line_4}
     \end{subfigure}
        \caption{Illustration of an occasional issue -  repeated line mask for line 1}
        \label{fig:Limitation}
\end{figure}

LineFormer is designed to allow multiple labels for the same pixels to address overlapping lines. This enables LineFormer to predict different masks for lines that have overlapping sections. However, occasionally this increases the likelihood that LineFormer produces duplicate masks for the same line. Fig [\ref{fig:Limitation}] shows a plot with four lines where LineFormer has predicted 5, a duplicate mask for line 1.

This issue can be mitigated by selecting the top-n predictions based on legend information and using Intersection Over Union (IoU) based suppression methods in its absence.

\section{Conclusion}
In this work, we have addressed the unique challenges that line charts pose in data extraction, which arise due to their structural complexities. Our proposed approach using Instance Segmentation has effectively addressed these challenges. Experimental results demonstrate that our method, LineFormer, achieves top-performing results across synthetic and real chart datasets, indicating its potential for real-world applications.
\\ \\ \\
\bibliographystyle{splncs04}
\bibliography{ref.bib} 
\end{document}